\pdfoutput=1

\documentclass[11pt]{article}

\usepackage[preprint]{acl}

\usepackage{times}
\usepackage{latexsym}
\usepackage{amsmath}
\usepackage[T1]{fontenc}

\usepackage[utf8]{inputenc}

\usepackage{microtype}

\usepackage{inconsolata}

\usepackage{graphicx}

\usepackage{url}

\usepackage{caption}
\usepackage{subcaption}
\usepackage{xspace}
\usepackage{algorithm}
\usepackage{algorithmicx}

\newcommand{\fig}[1]{\emph{Fig.~\ref{#1}}}
\newcommand{\sref}[1]{\emph{Section~\ref{#1}}}
\newcommand{\csrrag}{\emph{CSR-RAG}}
\newcommand{\tsql}{\emph{Text-to-SQL}}

\newif\ifnotes
\notesfalse

\newcommand{\rnote}[1]{\ifnotes $\ll$\textsf{\textcolor{red}{Rajpreet: {#1}}}$\gg$ \fi}

\newcommand{\nnote}[1]{\ifnotes $\ll$\textsf{\textcolor{blue}{Novak: {#1}}}$\gg$ \fi}

\title{CSR-RAG: An Efficient Retrieval System for Text-to-SQL on the Enterprise Scale}

\author{
  Rajpreet Singh$^{1}$,
  Novak Boškov$^{2}$,
  Lawrence Drabeck$^{2}$,
  Aditya Gudal$^{2}$,
  Manzoor A. Khan$^{2}$ \\
  $^{1}$Technical University of Munich, Germany \\
  $^{2}$Nokia Bell Labs \\
  \texttt{rajpreet.singh@tum.edu}, \\ \texttt{\{novak.boskov,lawrence.drabeck,manzoor.a.khan\}@nokia-bell-labs.com}, \\ \texttt{aditya.gudal@nokia.com}
}

\begin{document}
\maketitle
\begin{abstract}


Natural language to SQL translation (\tsql{}) is one of the long-standing problems that has recently benefited from advances in Large Language Models (LLMs).
While most academic \tsql{} benchmarks request schema description as a part of natural language input, enterprise-scale applications often require table retrieval before SQL query generation. To address this need, we propose a novel hybrid Retrieval Augmented Generation (RAG) system consisting of \textbf{\underline{c}}ontextual, \textbf{\underline{s}}tructural, and \textbf{\underline{r}}elational retrieval (\csrrag{}) to achieve computationally efficient yet sufficiently accurate retrieval for enterprise-scale databases.
Through extensive enterprise benchmarks, we demonstrate that \csrrag{} achieves up to 40\% precision and over 80\% recall while incurring a negligible average query generation latency of only 30ms on commodity data center hardware, which makes it appropriate for modern LLM-based enterprise-scale systems.



\end{abstract}

\section{Introduction}\label{sec:intro}

\nnote{The problem of baseline. Reviewer 3 is one to point that out.}
\nnote{Reviewer 3: do Google Scholar search for "value detection" in NL to SQL. Why just selecting the right tables is justifiable as a solution. ---> state of the art works only with BIRD and SPIDER.}
\rnote{This is an essential step fundamentally. lets discuss in detail }

Natural language-processing neural networks (NN) have recently shown significant advancements in generative tasks~\cite{radford2018improving,deepseek}, building upon the capabilities of the \emph{transformer} neural network architecture and the versatility of attention mechanisms~\cite{NIPS2017_3f5ee243}. Since their introduction, transformer-based NNs have continually improved their accuracy on tasks that involve large, well-defined evaluation sets, including a class of tasks commonly referred to as "verifiable"~\cite{drori2025diverseinferenceverificationadvanced}. One such amenable tasks is \tsql{}: given a natural language question, generate an SQL query that answers it~\cite{Katsogiannis-Meimarakis2023}. Although \tsql{} belongs to a wider class of code generation tasks~\cite{llmcodegen}, it involves several unique challenges, including the selection of the relevant database tables and columns to involve in the query (\emph{table retrieval}). 

\begin{figure}
    \centering
    \begin{subfigure}[b]{0.49\linewidth}
        \centering
    \includegraphics[width=\textwidth]{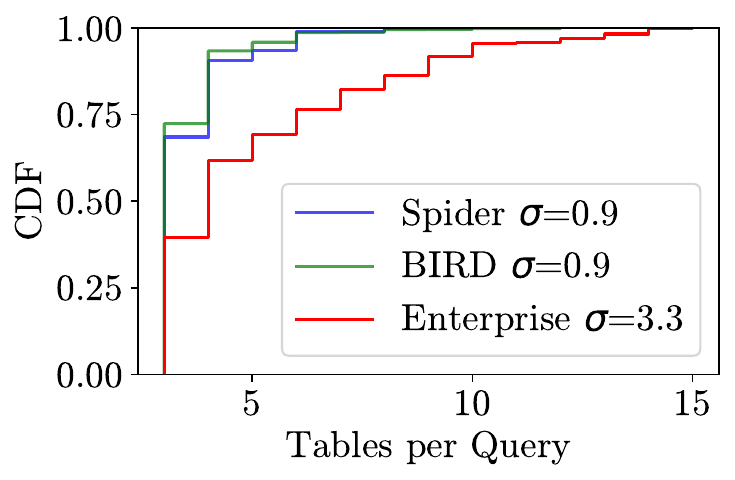}
    \end{subfigure}
    \begin{subfigure}[b]{0.49\linewidth}
        \centering
    \includegraphics[width=\textwidth]{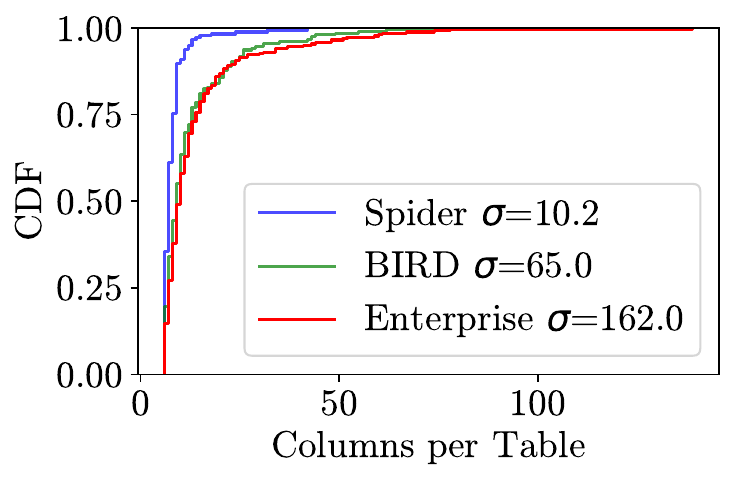}
    \end{subfigure}
    \begin{subfigure}[b]{\linewidth}
        \centering
    \includegraphics[width=\textwidth]{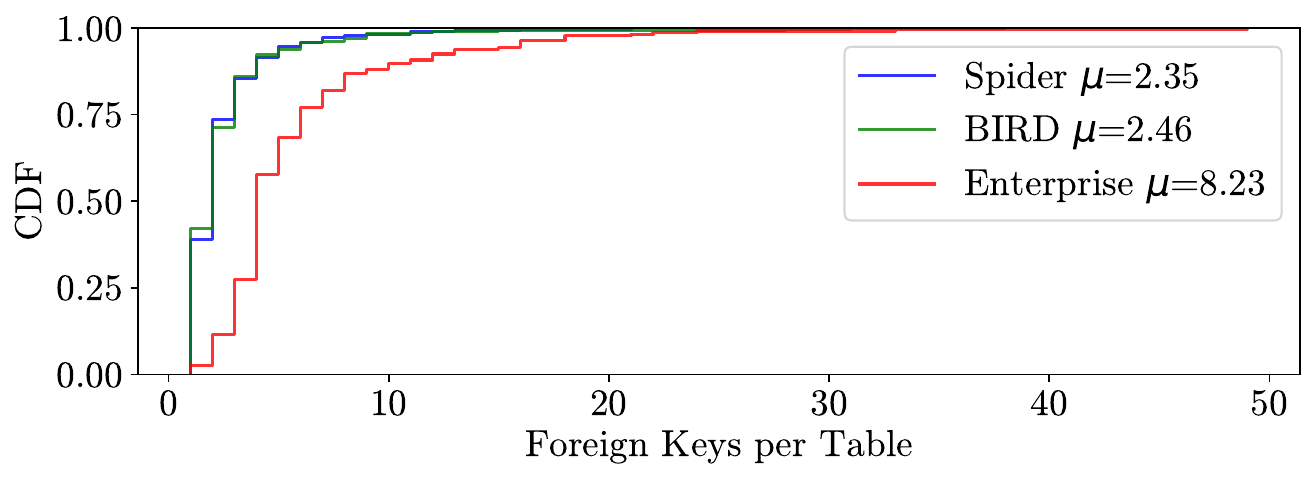}
    \end{subfigure}
    \caption{Cumulative Distribution Function (CDF) of an Enterprise Database Compared to Popular Benchmarks.}
    \label{fig:data_set}
\end{figure}

\nnote{Reviewer 3: "no proper description of the enterprise dataset": do a bit better dataset description job. Maybe a separate section.}
\rnote{Give simple description of nature of the database, maybe make a figure to compare spider/bird and enterprise.}


Given a natural language question $q$, we can define the problem of table retrieval as selecting the smallest set of table columns $\mathcal{T}$ (the \emph{relevant set}) that contains sufficient information to answer the question. While benchmark datasets such as Spider~\cite{spider} and BIRD~\cite{bird} provide the relevant set as an input alongside $q$, this approach is impractical in real-world enterprise settings, where the relevance set of an arbitrary question is typically unknown. Therefore, it is necessary to identify the relevant set from $q$ and pass it to the a language model-based generator, which then outputs the corresponding SQL query.


Besides the presupposition of the relevant set, we also observe significant statistical divergence between the popular benchmarks and typical enterprise databases, primarily with respect to the complexity of questions and database schema. As depicted in~\fig{fig:data_set}, queries in Spider and BIRD are significantly less likely to require large relevant sets compared to the \emph{Enterprise} case. While around 1/4 of \emph{Enterprise} queries necessitate 7 or more tables in their relevant set, this probability is negligible in Spider and BIRD, while their distributions also exhibit much smaller variance (0.9 versus 3.3). In other words, the popular benchmarks evaluate only a negligible amount of queries with many joins, the ones that often incur significant performance overheads. We observe a similar effect of missing out on "large queries" at the database schema level as well. While the median number of foreign keys per table is under 2.5 in Spider and BIRD, \emph{Enterprise} exhibits more than 3 times larger median with a non-negligible amount of tables with 10 or more foreign keys. 

When it comes to the size of tables with respect to the number of columns, BIRD in fact improves on top of Spider exhibiting a similar distribution to our \emph{Enterprise} case, except for the standard deviation that is noticeably larger in \emph{Enterprise}. Therefore, we believe that Spider and BIRD are insufficiently representative of enterprise-scale \tsql{} tasks. Considering the above factors, we present the following contributions:
\begin{itemize}
    \item \csrrag{}, a novel table retrieval system that simultaneously exploits semantic context of the question, the database schema, and its connectedness. Our two-stage process first executes contextual and graph retrieval in parallel, then combines their results using a hypergraph ranking subsystem.
    \item We evaluate the accuracy of \csrrag{} on an anonymized enterprise database achieving the precision of over 40\% and recall over 80\% for the most cases, satisfying the needs of modern LLM-based SQL query generators.
    \item Due to its parallelizable design, \csrrag{} achieves a superior query generation-time latency of around 30 milliseconds on commodity hardware.
\end{itemize}

The rest of this paper is organized as follows, in~\sref{sec:solution} we present the architecture of~\csrrag{} focusing on each individual component. In~\sref{sec:evaluation}, we evaluate the precision, recall, and latency of~\csrrag{}. In~\sref{sec:relatedwork}, we compare~\csrrag{} to \tsql{} approaches from the literature and conclude our paper in~\sref{sec:conclusion}.



\section{Solution Architecture}\label{sec:solution}

\begin{figure}
    \centering
    \includegraphics[width=0.5\textwidth]{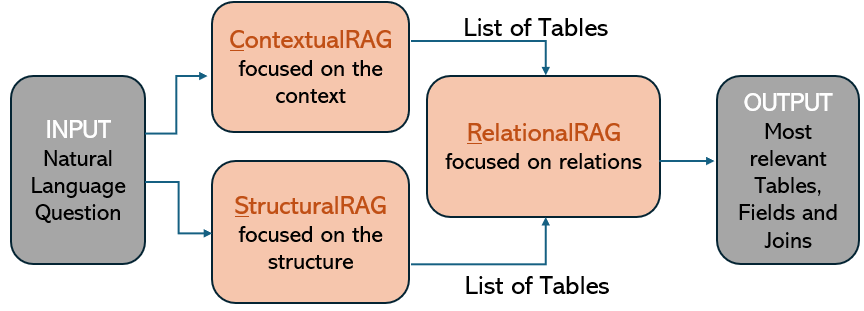}
    \caption{System Architecture.}
    \label{fig:architecture}
\end{figure}

\csrrag{}'s architecture comprises of three principal components: Contextual RAG, Structural RAG, and Relational RAG, organized as depicted in~\fig{fig:architecture} with the natural language question as the input and the list of most relevant table columns as the output supplied to the downstream LLM-based query generator. Contextual RAG incorporates natural language context following the intuition that contextually similar sentences yield similar SQL queries. Structural RAG turns database schema into its knowledge graph representation supplying the retriever with the knowledge of the database structure. At the end, Relational RAG captures the outputs of the parallel execution of Contextual and Structural RAGs and models multi-entity relationships within the narrowed table scope, producing the table columns that will participate in the joins of the final SQL query.



\subsection{Contextual RAG}

\begin{figure}
    \centering
    \includegraphics[width=0.5\textwidth]{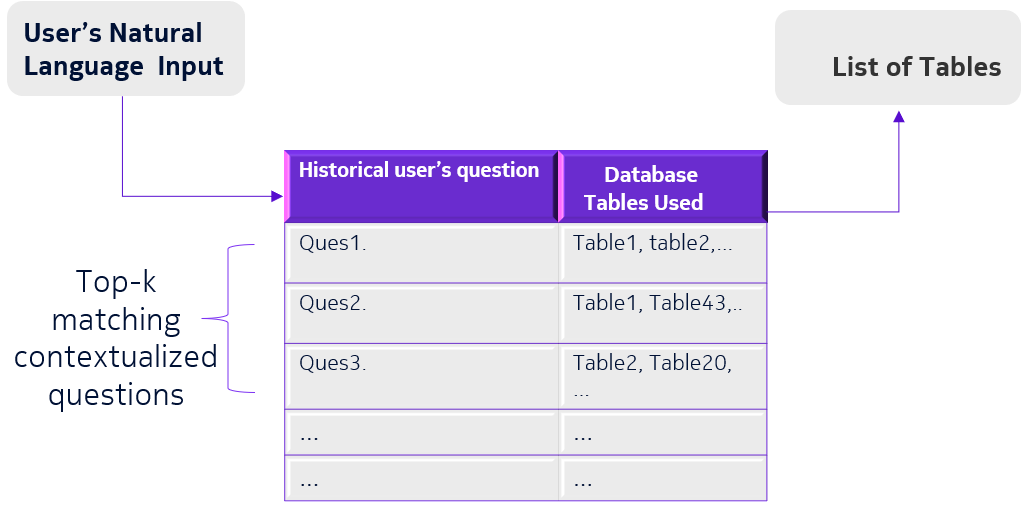}
    \caption{The Function of Contextual RAG.} 
    \label{fig:contextual}
\end{figure}

Capturing the context of a natural language questions is crucial in measuring its similarity to the others in the ground truth data set. \nnote{Add a sentence to explain why the measure of similarity is needed.} \rnote{Again, a fundamental step in retrieval systems. To compare the input query to whatever context we have around the prior knowledge} To allow for such a function, our Contextual RAG first performs sentence embedding using a BERT-like encoder~\cite{bert}, mapping the sentence into a multi-dimensional vector space. By doing so, we avoid the semantic isolation drawbacks of some traditional RAGs. \nnote{"BERT-like encoder helps avoid semantic isolation drawbacks of some traditional RAGs."} As in~\fig{fig:contextual}, Contextual RAG then retrieves a list of $k$ top similar questions from the ground truth data set and associates the retrieved similar questions to their relevant sets (tables used in the query), then returns the union of all $k$ relevant sets. \nnote{addressing the Reviewer 3: either omit contextual RAG as a component or redefine the context not to be the ground truth set but the table descriptions or something else.}


Our Contextual RAG enables global search across diverse database schemas, pinpointing domain-specific structures without requiring explicit identification. It requires two tuning parameters, the number of chunks to be retrieved and the quality of contextualization, which we need to balance to allow for comprehensive retrieval while retaining computational efficiency. Through our experiments, we observed that simpler similarity metrics such as cosine similarity yield comparable results to more complex ones such as BM25~\cite{SPARCKJONES2000779}.

Let \( Q \) be the set of all natural language queries, and let \( \mathcal{D} \) represent the database schema consisting of \( T \) tables and \( F \) fields. Each SQL query \( S \) corresponds to a natural language question \( q \in Q \), forming a chunk $C_i = (q_i, S_i)$, for $i \in \{1, 2, \dots, N\}$, where $N$ is the count of retrieval units in the ground truth dataset. We define the contextualization function \( \phi \) that enhances a query by incorporating schema-related information as:
\nnote{Reviewer 3: Is D the input to our RAG? No, we don't use it in retrieval time, but only when we are contextualizing the ground-truth dataset.}
\[
\phi(q) = q + \text{desc}(S, \mathcal{D}),
\]
where \( \text{desc}(S, \mathcal{D}) \) includes table names, column names, and their textual descriptions. We then use our similarity metric \( Sim \) and tuning parameter \( k \) to retrieve the similar chunks as:
\[
\mathcal{R}(q') = \operatorname{argmax}_{C_i} Sim(q', C_i), \quad | \mathcal{R}(q') | = k.
\]
The larger \( k \), the more tables retrieved and the larger the probability of retrieving unnecessary information. Therefore, we set \( k \) to optimize for precision and recall, with more details in~\sref{sec:evaluation}.

\subsection{Structural RAG}

\begin{figure}[htbp] 
    \centering
    \includegraphics[width=0.5\textwidth]{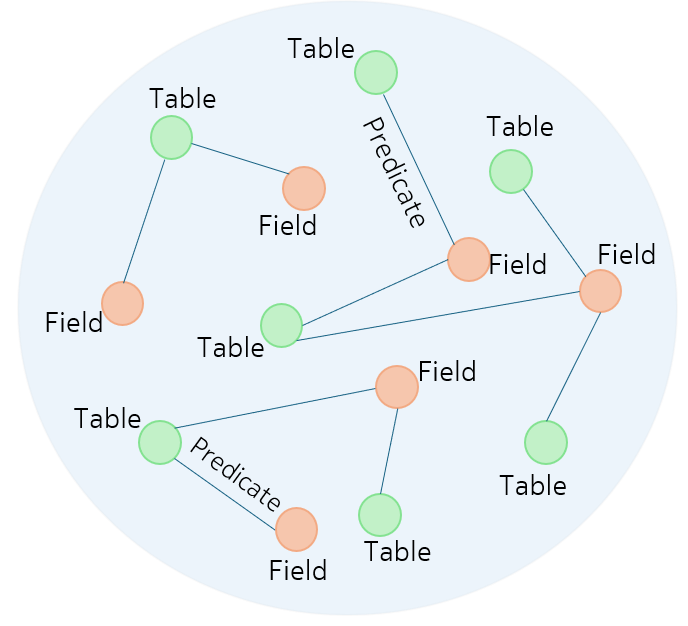}
    \caption{Graph Representation of the Schema.}
    \label{fig:graph_schema}
\end{figure}

Effective representation of prior knowledge is paramount in retrieval systems, potentially offering substantial accuracy gains and the knowledge graphs emerged as a de-facto standard in this area. While conventional approaches use language models to generate graph triplets <subject, predicate, object>, they suffer from inconsistent predicate inference. To overcome this drawback and keep the retrieval latency low, we opt for a deterministic approach to building the knowledge graph from database schema. Our graph representation consists of <field, "Relation", table> triplets, where "Relation" denotes "is a column of" (see~\fig{fig:graph_schema}). Therefore, the total number of triplets is equal to the number of tables times the number of columns in the schema, optimizing for representation accuracy in representing enterprise-scale databases.


Let the database schema have a set of tables $T$ and fields $F$. The knowledge representation is then a graph \(G = (V, E)\), where \(V = T \cup F\) and \(E \subseteq V \times V\).
Each triplet in the knowledge graph is represented as \((f_i, t_j) \in F \times T\), where $f_i \in F$ is a column and $t_j \in T$ is a table.
Given a natural language question $q$ and a tuning parameter $l$, the retrieval function $\gamma$ returns the top-$l$ triplets:
\[
\gamma(q, G, l) \rightarrow \{(f_i, t_j)_1, ..., (f_i, t_j)_l\},
\]
where the tuning parameter \(l\) controls the number of retrieved triplets from the knowledge graph.
Similar to Contextual RAG, a large \( l \) retrieves more tables.







\subsection{Relational RAG}

\begin{figure}[htbp] 
    \centering
    \includegraphics[width=0.5\textwidth]{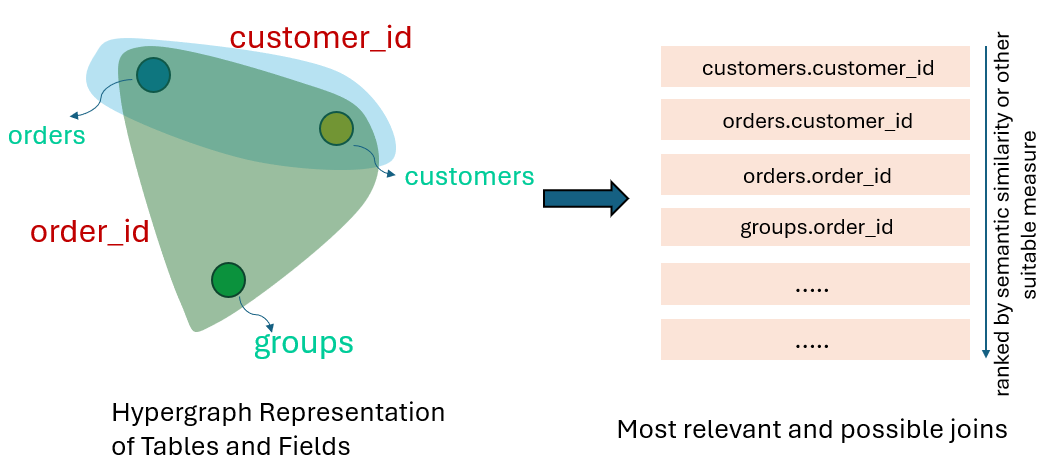}
    \caption{Hypergraph Ranking to Select the Most Relevant Scoped Columns.} 
    \label{fig:hypergraph_rep}
\end{figure}

The final component of the system addresses a critical aspect of database operations: foreign key relationships, which are essential for generating accurate SQL queries when leveraging fine-tuned LLMs. The Relational RAG component captures relationships among tables retrieved by Contextual and and Structural RAGs, ensuring that the table joins are contextually relevant to the natural language query. Unlike the simpler knowledge graph representation in Structural RAG, Relational RAG  encodes its knowledge into a hypergraph to handle the increased complexity, as shown in~\fig{fig:hypergraph_rep}. We represent tables as the nodes in the hypergraph and columns as its hyperedges. Nodes are connected by a hyperedge if the column encoded by the hyperedge is likely to be joined on in an SQL query. The columns that are likely to join several tables are encoded as the hyperedges with many nodes.
This modeling principle is particularly suitable for database schemas with complex foreign-key relationships such as our \emph{Enterprise} example from~\sref{sec:intro}, as it allows for flexible modeling of multi-table dependencies.
The function of Relational RAG therefore consists of: 1) dynamic hypergraph construction from narrowed table set outputted by Contextual and Structural RAGs, 2) capturing the semantics of SQL joins, and 3) preparing the \texttt{table.column} format for the fine-tuned LLM query generator.


\begin{algorithm}[H]
\caption{Hypergraph Ranking.}
\label{alg:a1}
\begin{algorithmic}[1]
    \State $h$ : Number of nodes to select
    \State $N$ : Set of nodes
    \State $T$ : Task metadata requirements
    \State $O$ : Operation functions for scoring
    \State $W$ : Weight factors for different metrics


\State \textbf{func} HypergraphRanking ({$h$, $N,$ $T$, $O$, $W$})
    
    \State $scoredNodes \gets \emptyset$
    
    \State {\textbf{for all} node $v \in$ $N$ \& hyperedge $e \in$ $T$}
        \State {      } $M_v \gets$ metadata of node $v$
        \State {      } \textbf{if} {$M_v$.availability = false} \textbf{then}
            \State {      }  {      } \textbf{continue}
        \State  {      } \textbf{end if}

        \State  {      } $(v \otimes e ) \gets \textbf{compute with operator}; \otimes \in O$
        \State  {      } \textbf{if }{$w_v > 0 \textbf{  where  } w_v \in W$ \textbf{then}}
            \State  {      }  {      } $score_{v,e} \gets (v \otimes e ) / w_v$
        \State   {      } \textbf{else}
            \State  {      } {      } $score_{v,e} \gets 0$
        \State {      } \textbf{end if}
        
        \State  {      }  {      }$scoredNodes \gets scoredNodes \cup   {      }\{(v, score_{v,e})\}$
    \State \textbf{end for}
    
    \State \textbf{SORT} $scoredNodes$ by score in descending order
    \State $selectedNodes \gets$ First $h$ nodes from $scoredNodes$
    
    \State \textbf{RETURN} $selectedNodes$
\end{algorithmic}
\end{algorithm}

The core of Relational RAG is our general operator-based hypergraph ranking algorithm (see \emph{Algorithm~\ref{alg:a1}}), which relies on ranking hyperedge-node concatenations to identify potential joins. Similarly to the previous RAG components, Relational RAG makes use of a similarity metric to pick the top-$k$ hyperedge-node pairs, effectively prioritizing the most relevant valid table joins based on the context of the input questions. By integrating contextual and structural information from previous components and leveraging the hypergraph ranking, Relational RAG ensures sufficiently accurate yet efficient retrieval of table-join information, enhancing SQL generation for complex queries (see~\fig{fig:hypergraph_rep}).

Let $H = (V, E)$ be a hypergraph where $V$ is the set of tables retrieved from Contextual and Structural RAGs, $E$ are the hyperedge representations of columns, and $\Upsilon(v, e)$ is the relevance score of the semantic entity ($v \otimes e$) in the context of the natural language question $q$, where $\otimes$ denotes concatenation of a table and its column. The result of Relational RAG is then output of the similarity ranking:
\[
\beta(q, {H}) \rightarrow \{(v_1 \otimes e_1), (v_2 \otimes e_2), ..., (v_k \otimes e_h)\},
\]
where \((e_i, v_i)\) are ranked and ordered from the entire hypergraph representation. The ranking function \(\beta\) performs a similarity measure between the question \(q\) and the hypergraph \(H\) using $\Upsilon(v,e)$, which we implement as cosine similarity of the semantic entity \((v \otimes e)\) and question \(q\).
The tuning parameter \(h\) defines the number of semantic entities to be retrieved by the Relational RAG component. Note that the hypergraph ranking algorithm can take various forms depending on the semantic meaning attached to $\otimes$. For instance, one can add more context around this operator to involve tables and column descriptions, which would further enhance the retrieval accuracy.


\section{Experimental Evaluation}~\label{sec:evaluation}

\begin{table}
    \centering
    \begin{tabular}{||c c c||} 
 \hline
 Group & Tables & Columns \\ [0.5ex] 
 \hline\hline
 1 & 50 & 701 \\ 
 2 & 100 & 1486  \\
 3 & 200 & 2567 \\
 4 & 246 & 3021 \\
 \hline
\end{tabular}
    \caption{The Sizes of Experimental Table Groups in the ~\emph{Enterprise} dataset.}
    \label{tab:groups_sizes}
\end{table}

In this section, we evaluate \csrrag{} on our anonymized \emph{Enterprise} database from~\sref{sec:intro}. We consider four groups of database schemas and their related ground truth traces consisting of natural language questions and corresponding SQL queries. The number of tables and columns per group are given in \emph{Table}~\ref{tab:groups_sizes}. In our evaluations, we focus on iterative implementations of Contextual, Structural, and Relational RAGs, where iterations represent recurrent passes through the RAG component. For instance, the second iteration of Contextual RAG takes the output of the first iteration of Contextual RAG alongside a decreased tuning parameter. If not otherwise noted, we write the hyperparameter values in each iteration as the numbers above the data point in our plots.

\nnote{Reviewer 3: better job at describing hyper parameters.}
\rnote{Methodology to come at the right combination of hyper-parameters.}
\begin{figure}
    \centering
    \includegraphics[width=0.5\textwidth]{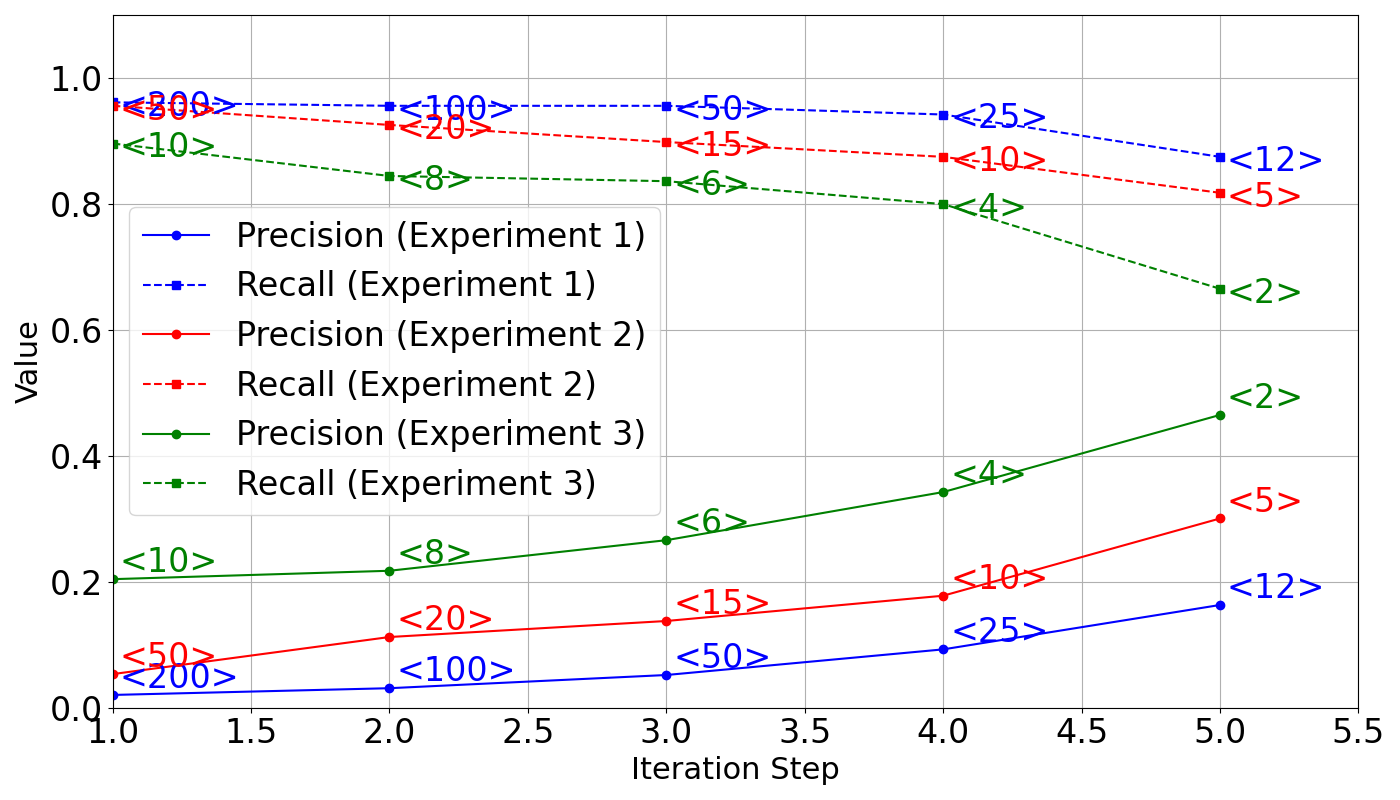}
    \caption{Precision and Recall for Contextual RAG on Group 2.}
    \label{fig:cr1}
\end{figure}

In \fig{fig:cr1}, we plot the precision and recall of Contextual RAG as a function of the number of iterations. As apparent from the bond of solid lines in~\fig{fig:cr1}, a decreasing chunk size \(k\) leads to an increased Contextual RAG precision. This trend is expected due to a reduction in the search space across the iterations, defining the search space as the space of all possible tables in the experimental table group. However, we also see a moderate decrease in recall attributed to an increased number of false negatives, which is also the consequence of the reduced number of selected tables.

As shown in~\fig{fig:cr2}, the initial choice of \(k\) only moderately affects precision. \nnote{Clarify for Reviewer 3. The INITIAL choice of k does not matter.}The choice of a very small \(k\) in a single iteration does not lead to a high precision hike due to the insufficient contextual information being retrieved, resulting in potentially incomplete or inaccurate answers. The loss of relevant contextual information is high in a single step retrieval with small \(k\), for instance 2 or 4.
\begin{figure}
    \centering
    \includegraphics[width=0.5\textwidth]{latex/plot_CR_exp35.pdf}
    \caption{Precision and Recall for Contextual RAG on Group 2.}
    \label{fig:cr2}
\end{figure}

\begin{figure}
    \centering
    \includegraphics[width=0.5\textwidth]{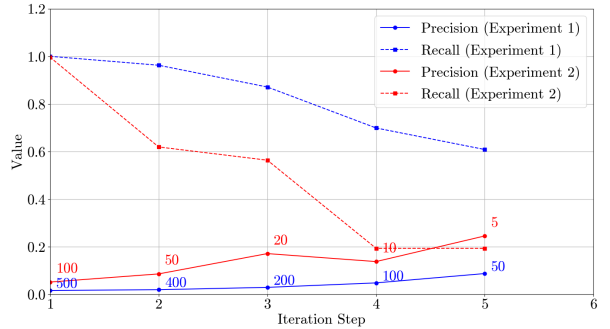}
    \caption{Precision and Recall for Structural RAG on Group 2.}
    \label{fig:gr1}
\end{figure}

Similarly to Contextual RAG, the precision and recall for Structural RAG are shown in \fig{fig:gr1}. The trend is similar to that of Contextual RAG, but there is a sharp drop in recall due to lack of contextual information in the knowledge graph representation of the database schema. The tuning parameter \(l\), the size of retrieved triplets from the knowledge graph are  highlighted at the data points in \fig{fig:gr1}.

\nnote{Reviewer 3: better discussion on how to choose the optimal k balancing precision and recall.}
\rnote{Optimal balance is highly subjective to the quality of prior knowledge. And also how complex the knowledge graph.}

\begin{figure}
    \centering
    \includegraphics[width=0.5\textwidth]{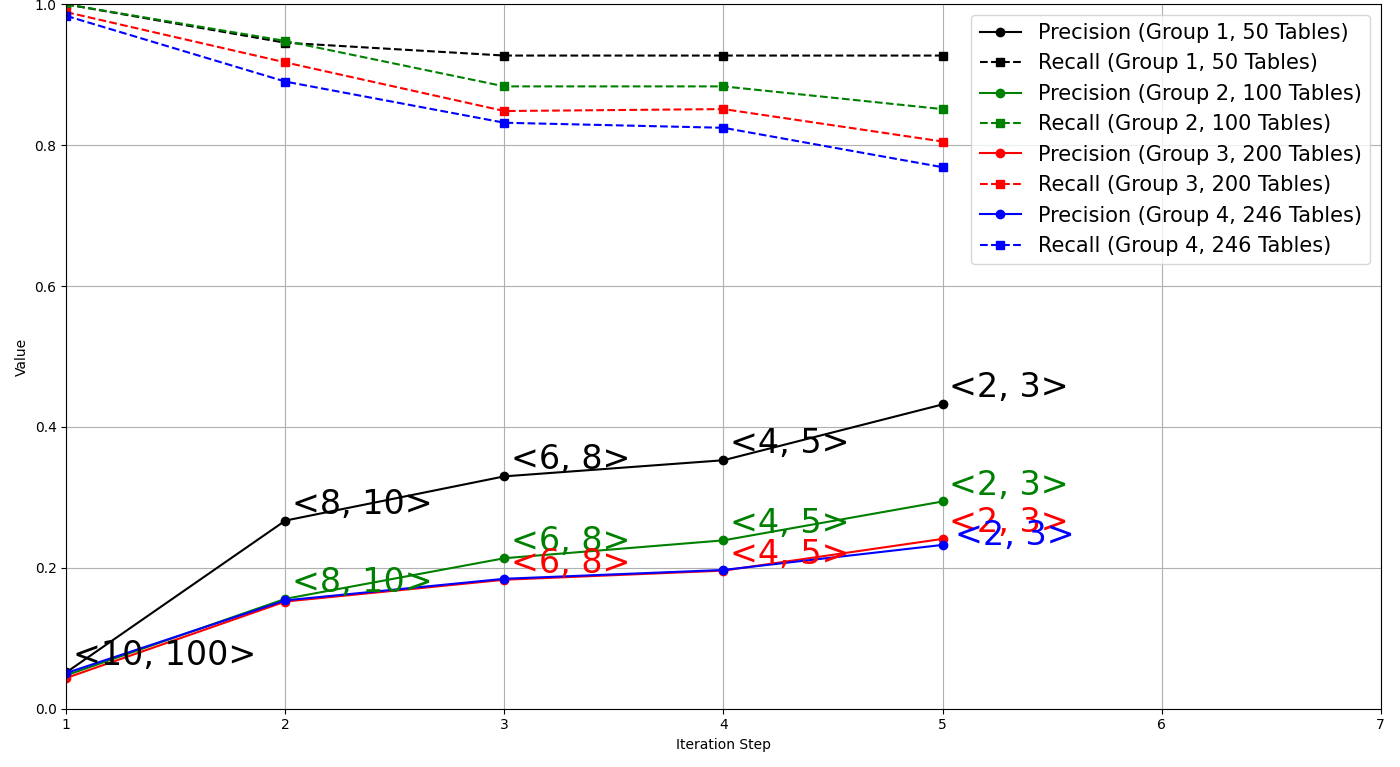}
    \caption{Precision and Recall for Contextual and Structural RAG Components Combined for all Groups.}
    \label{fig:cr_gr1}
\end{figure}

\begin{figure}[htbp] 
    \centering
    \includegraphics[width=0.5\textwidth]{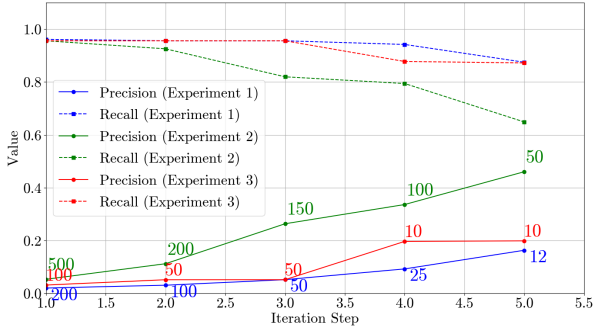}
    \caption{Precision and Recall for the Relational RAG Component on Group 3.}
    \label{fig:rg1}
\end{figure}

A combined evaluation of Contextual and Structural RAG is shown in \fig{fig:cr_gr1}. For each group, the best configuration of tuning parameters \((k, l)\) is highlighted at each iteration step. The choice of these parameters might differ among ground truth data sets,  but we do not expect significant changes in the trend.
After the retrieval of tables using Contextual, and Structural RAG, our system performs ranking using hypergraph representation of selected tables to retrieve most relevant columns that are necessary for generating efficient joins. Following the same conventions from the previous plots, we plot the precision and recall results of Relational RAG in~\fig{fig:rg1}.

\begin{figure}
    \centering
    \includegraphics[width=0.5\textwidth]{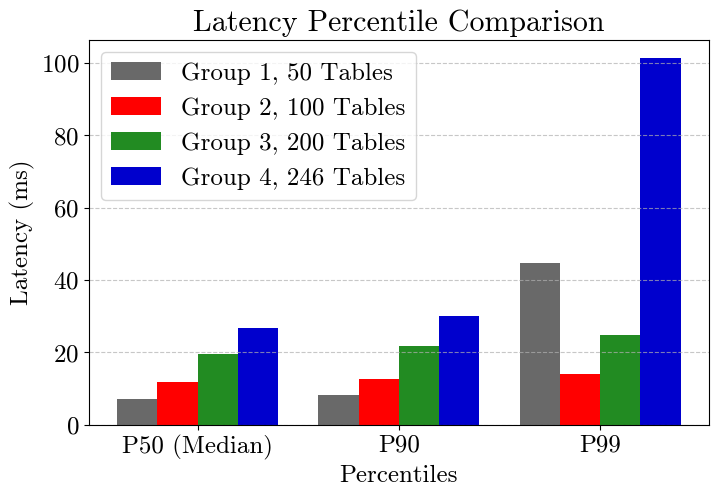}
    \caption{Overall Latency of \csrrag{}.}
    \label{fig:latency}
\end{figure}

Finally, we measure the incurred latency of entire \csrrag{} on all groups and plot the results in~\fig{fig:latency} reporting on the 50th, 90th, and 99th percentile. Our results suggest that \csrrag{} takes around 30ms on average to perform the full table retrieval task, which is negligible compared to the time the state-of-the art LLM-based query generators take to generate the SQL queries. We obtain the latency results from~\fig{fig:latency} on Intel Xeon Gold 5317 with 3GHz cores, a relatively common datacenter CPU.

\section{Related Work}~\label{sec:relatedwork}

\cite{huang2022mixedmodalityrepresentationlearningpretraining} propose an optimized dense retriever for joint table-text evidences for OpenQA. \cite{chen2025tableretrievalsolvedproblem} highlight the "table-retrieval" problem in \tsql{} and propose a re-ranking method based on mixed integer programming to select the best set of tables based on joint decisions about relevance and compatibility. As opposed to these works, we not only retrieve tables, but also the columns involved in potential joins, thus optimizing for complex large-scale databases. We also do it in a computationally tractable and determinist fashion as opposed to non-deterministic retrieval systems such as \cite{li2023llmservedatabaseinterface}, \cite{mao-etal-2024-enhancing}, \cite{li-etal-2024-multisql}, \cite{wang-etal-2023-know}, while our knowledge graph-based schema representation enables for more accurate question answering as suggested by~\cite{10.1145/3661304.3661901}. On the other hand, a recent work on hybrid retrieval system called HYBGRAG~\cite{lee2024hybgraghybridretrievalaugmentedgeneration} takes an "agentic" approach and therefore suffers from prohibitive latency overheads. Another work based on a classification based table selection~\cite{Chopra2024EnhancingNL} offers a practical solution for improving the accuracy and efficiency of \tsql{}~\cite{sun-etal-2023-exploratory}, \cite{jiao-etal-2024-text2db}, \cite{xiang-etal-2023-g} models, but requires a huge training dataset and struggles to scale for enterprise databases. Noteworthy is~\cite{hu-etal-2023-importance} that offers a synthesis framework to generate high-quality data for \tsql{} due to the lack of public datasets from enterprise databases.

\section{Conclusions}~\label{sec:conclusion}
In this work, we presented~\csrrag{}, a system that addresses a critical challenge in information retrieval for \tsql{}. In ~\csrrag{}, we divide the retrieval problem into three inherent aspects: the semantic context of natural language questions, database structure, and connectedness among the database tables. We achieve a sufficiently high precision and recall for enterprise-scale databases combining the output from Contextual and Structural RAGs within our hypergraph ranking-based Relational RAG. Due to ~\csrrag{}'s superior latency over similar approaches from the literature, we use ~\csrrag{} as a real-time inferencing component in front of a state-of-the-art LLM-based SQL query generator, achieving industry-grade SQL generation accuracy and an exceptional run-time performance.       










\bibliography{custom}



\end{document}